 \documentclass[sigconf]{acmart}
\usepackage{multirow}
\usepackage{subcaption}
\usepackage{placeins}

\usepackage{enumitem}
\usepackage{balance}

\AtBeginDocument{%
  }

\copyrightyear{2026}
\acmYear{2026}
\setcopyright{cc}
\setcctype{by}
\acmConference[GECCO '26]{Genetic and Evolutionary Computation Conference}{July 13--17, 2026}{San Jose, Costa Rica}
\acmBooktitle{Genetic and Evolutionary Computation Conference (GECCO '26), July 13--17, 2026, San Jose, Costa Rica}
\acmDOI{10.1145/3795095.3805148}
\acmISBN{979-8-4007-2487-9/2026/07}

\begin{document}


\title[The Role of Teacher-Student Smoothness Alignment in Genetic Programming-based Symbolic Distillation]{Teaching the Teacher: The Role of Teacher-Student Smoothness Alignment in Genetic Programming-based Symbolic Distillation}


\author{Soumyadeep Dhar}
\email{sdhar1602@kgpian.iitkgp.ac.in}
\affiliation{%
\institution{Indian Institute of Technology Kharagpur}
\department{Department of Industrial and Systems Engineering}
}

\author{Kei Sen Fong}
\email{fongkeisen@u.nus.edu}
\orcid{0009-0000-4135-4858}
\affiliation{%
 \institution{National University of Singapore}
 \department{Department of Electrical and Computer Engineering}
}

\author{Mehul Motani}
\email{motani@nus.edu.sg}
\orcid{0000-0003-3262-0207}
\affiliation{%
 \institution{National University of Singapore}
 \department{Department of Electrical and Computer Engineering, Institute of Data Science, N.1 Institute for Health, Institute for Digital Medicine}
}


\begin{abstract}
Obtaining human-readable symbolic formulas via genetic program\-ming-based symbolic distillation of a deep neural network trained on the target dataset presents a promising yet underexplored path towards explainable artificial intelligence (XAI); however, the standard pipeline frequently yields symbolic models with poor predictive accuracy. We identify a fundamental misalignment in functional complexity as the primary barrier to achieving better accuracy: standard Artificial Neural Networks (ANNs) often learn accurate but highly irregular functions, while Symbolic Regression typically prioritizes parsimony, often resulting in a much simpler class of models that are unable to sufficiently distill or learn from the ANN teacher. To bridge this gap, we propose a framework that actively regularizes the teacher's functional smoothness using Jacobian and Lipschitz penalties, aiming to distill better student models than the standard pipeline. We characterize the trade-off between predictive accuracy and functional complexity through a robust study involving 20 datasets and 50 independent trials. Our results demonstrate that students distilled from smoothness-regularized teachers achieve statistically significant improvements in $R^{2}$ scores, compared to the standard pipeline. We also perform ablation studies on the student model algorithm. Our findings suggest that smoothness alignment between teacher and student models is a critical factor for symbolic distillation.

\end{abstract}

\ccsdesc[500]{Computing methodologies~Genetic programming}
\ccsdesc[500]{Computing methodologies~Regularization}
\ccsdesc[500]{Computing methodologies~Neural networks}

\keywords{Knowledge Distillation, Symbolic Regression, Explainable AI, Smoothness Regularization}


\maketitle

\section{Introduction}
\label{sec:intro}

The remarkable success of deep learning has revolutionized fields from computer vision to natural language processing~\cite{lecun2015deep}. These models, particularly deep artificial neural networks (ANNs), have achieved state-of-the-art performance on a vast array of predictive tasks. However, this performance often comes at the cost of transparency. ANNs operate as ``black boxes,'' with their immense number of parameters forming complex, non-linear relationships that are opaque to human understanding~\cite{guidotti2018survey}. This lack of interpretability poses a significant barrier to their adoption in high-stakes domains such as scientific discovery~\cite{lipton2017mythosmodelinterpretability, 13cranmer2023interpretablemachinelearningscience, 15petersen2021deepsymbolicregressionrecovering, 16udrescu2020aifeynman20paretooptimal, bartlett2025introductionsymbolicregressionphysical}, medical diagnostics~\cite{topol2019high, rudin2019stop, fong2024explainable, fong2024symbolic}, and financial modeling, where the ability to trust, debug, and understand a model's reasoning is as critical as its predictive accuracy \cite{jobin2019global}.

To bridge this gap, the field of explainable artificial intelligence (XAI) has explored various methods to shed light on these models. One promising paradigm is knowledge distillation~\cite{hinton2015distilling}, where a large, complex ``teacher'' model transfers its knowledge to a smaller, simpler ``student'' model. While often used for model compression, this framework also provides a powerful pathway to interpretability by selecting a student model that is inherently transparent~\cite{craven1996extracting}.

Among possible student models, those produced by Symbolic Regression (SR) are particularly compelling, as SR aims to discover the underlying mathematical formula that maps inputs to outputs~\cite{koza1992genetic, q10.1145/3735634}. The result is not an explanation, but the model \textit{itself}: a fully transparent equation. Recent work has attempted to refine or replace components of trained neural networks using symbolic regression \cite{wei2025refining} However, the naive pipeline of training an ANN and then having an SR model mimic it often fails on real-world data \cite{14lacava2021contemporarysymbolicregressionmethods, 19virgolin2022symbolicregressionnphard}, resulting in low-fidelity symbolic models (as shown in Section~\ref{sec:experiments}). This suggests a fundamental problem: the standard teacher network is a poor instructor.

However, the promise of this naive pipeline is seldom realized in practice. We find that this is due to a fundamental misalignment in functional complexity: standard ANNs, while accurate, learn ``spiky'' functions with high Jacobian and Lipschitz norms, making them poor teachers. Conversely, directly using SR produces simple, smooth functions but is often highly inaccurate. Instead of passively accepting this trade-off, we ask: \textit{can we train a teacher to be both accurate and simple?} In this work, we investigate the use of functional smoothness regularizers (both Jacobian and Lipschitz) to find a ``sweet spot'' teacher. We are the first, to the best of our knowledge, to quantitatively characterize the full trade-off between a teacher model's accuracy ($R^2$) and its functional smoothness (e.g., Avg.\ Jac.\ Norm).


The main contributions of this paper are as follows:
\begin{enumerate}[nosep,  wide=0pt]
\item We quantitatively demonstrate the core challenge: standard ANNs are accurate but spiky (high-$R^2$, high Jac. Norm), while directly using SR (i.e., \texttt{direct\_sr}) is smooth but inaccurate (low-$R^2$, low Jac. Norm).
\item We show that by tuning the strength ($\lambda$) of functional smoothness regularizers (Jacobian or Lipschitz), we can create a ``sweet spot'' teacher model that is both significantly more accurate and functionally smoother than the \texttt{direct\_sr} baseline.
\item We validate this framework through a comprehensive empirical study across 20 diverse regression datasets with 50 independent trials per dataset. Global statistical analysis using the Wilcoxon Signed-Rank Test confirms that these optimized teachers are the key to successful distillation, yielding student models with significantly higher $R^{2}$ scores ($p < 10^{-6}$) compared to standard baselines.
\end{enumerate}

\section{Related Work}
\label{related}

Our research is situated at the intersection of several key areas in machine learning: knowledge distillation, explainable AI, symbolic regression, and network regularization.

\subsection{Knowledge Distillation}
First popularized by Hinton et al.~\cite{hinton2015distilling}, knowledge distillation is a technique where a compact ``student'' model is trained to reproduce the behavior of a larger, pre-trained ``teacher'' network. The original motivation was primarily model compression, enabling the deployment of powerful models on resource-constrained devices. However, the paradigm is more general \cite{22stanton2021doesknowledgedistillationreally, 24Gou_2021} and has been adapted for tasks such as transferring knowledge across different data modalities and for semi-supervised learning. Our work reframes distillation's purpose, using it not for compression, but as a mechanism to transfer knowledge from a high-performance but opaque model to one that is inherently interpretable.

\subsection{Explainable AI (XAI) and Interpretable Models}
The challenge of understanding black box models has given rise to the field of XAI. Current approaches can be broadly categorized. \textit{Post-hoc} explanation methods, such as LIME~\cite{ribeiro2016lime} and SHAP~\cite{lundberg2017unified}, aim to explain individual predictions of a pre-trained model without altering it. While powerful, these methods provide local explanations and do not reveal the model's global logic. In contrast, \textit{inherently interpretable models} are transparent by design. This category includes linear models, decision trees, and Generalized Additive Models (GAMs). Our work aligns with the goal of creating inherently interpretable models. However, instead of training a simple model directly on the data (which may yield suboptimal performance), we aim to extract a simple model from a high-performance complex teacher, seeking to combine the predictive power of deep learning with the transparency of symbolic equations.

\subsection{Symbolic Regression}
At the core of our student model is Symbolic Regression, a machine learning technique that searches the space of mathematical expressions to find the function that best fits a given dataset. Typically implemented using Genetic Programming (GP) as pioneered by Koza~\cite{koza1992genetic}, SR has been a powerful tool for automated scientific discovery, rediscovering physical laws from experimental data. Beyond traditional evolutionary methods, recent advancements have integrated Deep Reinforcement Learning to optimize the search for mathematical expressions \cite{15petersen2021deepsymbolicregressionrecovering, 30NEURIPS2021_d073bb8d}, or exploited graph modularity to accelerate scientific discovery \cite{16udrescu2020aifeynman20paretooptimal}. While SR is traditionally applied directly to data, we re-purpose it as the ``student'' in our distillation pipeline, tasking it with a novel goal: to discover the mathematical function implicitly learned by a neural network.

Recent work has explored distilling specialized (e.g., graph network message approximation) models into symbolic or feature-based representations using carefully designed inductive biases \cite{cranmer2020discovering, fongfeat}. In contrast, our work focuses on general regression and also explicitly controlling the functional smoothness of the teacher model itself.

\subsection{Regularization for Functional Smoothness}
Regularization is a cornerstone of training deep neural networks. While techniques like L1/L2 weight decay and Dropout~\cite{srivastava2014dropout} constrain the parameter space to prevent overfitting, a distinct class of regularizers aims to control the complexity of the \textit{function space} by encouraging smoothness.

Penalizing the norm of the network's Jacobian is a well established technique for this purpose, which we repurpose from its prior use in improving adversarial robustness~\cite{hoffman2019robust} for the novel objective of distillability. An alternative approach is to enforce a global smoothness constraint by regularizing the network's Lipschitz constant, often approximated by penalizing the spectral norm of the weight matrices~\cite{gouk2021regularisation, 26bartlett2017spectrallynormalizedmarginboundsneural, 2810.1145/3446776}. Our work is the first to investigate and compare these two approaches to determine which type of smoothness, local or global, is more effective for making a neural network a better teacher for a symbolic student.

\section{Methodology}
\label{sec:methodology}

In this section, we formally define the baseline ANN-to-Symbolic-Regression pipeline and then describe the two functional smoothness regularizers: Jacobian and Lipschitz, that we investigate.

\subsection{The ANN-to-SR Distillation Pipeline}
Our approach builds upon a standard two-stage pipeline for distilling a trained neural network into a symbolic formula.

\begin{table*}[t]
\centering
\caption{Effect of Smoothness Regularization on Neural Network Stability and Accuracy}
\label{tab:combined_results}

\begin{subtable}{\textwidth}
\centering
\caption{Average Jacobian Norm}
\small  
\resizebox{\textwidth}{!}{
\begin{tabular}{l|ccccccccccccc}
\hline
Dataset & \texttt{direct\_sr} & \texttt{vANN} & \texttt{l2} & \texttt{lip\_0.1} & \texttt{lip\_0.3} & \texttt{lip\_0.5} & \texttt{lip\_0.7} & \texttt{lip\_1} & \texttt{jac\_0.1} & \texttt{jac\_0.3} & \texttt{jac\_0.5} & \texttt{jac\_0.7} & \texttt{jac\_1} \\
\hline
California Housing & 0.630 & 34.94 & 31.62 & 1.226 & 0.429 & 0.244 & 0.147 & 0.049 & 0.708 & 0.392 & 0.262 & 0.183 & 0.142 \\
Concrete Strength & 1.856 & 4.169 & 3.807 & 1.348 & 0.551 & 0.278 & 0.157 & 0.053 & 1.027 & 0.434 & 0.286 & 0.215 & 0.151 \\
\hline
\end{tabular}
}
\end{subtable}

\vspace{0.5\baselineskip}  

\begin{subtable}{\textwidth}
\centering
\caption{Lipschitz Approximation}
\small
\resizebox{\textwidth}{!}{
\begin{tabular}{l|ccccccccccccc}
\hline
Dataset & \texttt{direct\_sr} & \texttt{vANN} & \texttt{l2} & \texttt{lip\_0.1} & \texttt{lip\_0.3} & \texttt{lip\_0.5} & \texttt{lip\_0.7} & \texttt{lip\_1} & \texttt{jac\_0.1} & \texttt{jac\_0.3} & \texttt{jac\_0.5} & \texttt{jac\_0.7} & \texttt{jac\_1} \\
\hline
California Housing & 0.794 & 12.52 & 11.54 & 1.191 & 0.683 & 0.511 & 0.392 & 0.224 & 1.087 & 0.706 & 0.582 & 0.496 & 0.490 \\
Concrete Strength & 3.695 & 2.924 & 2.772 & 1.416 & 0.844 & 0.592 & 0.434 & 0.266 & 1.295 & 0.868 & 0.765 & 0.654 & 0.556 \\
\hline
\end{tabular}
}
\end{subtable}

\vspace{0.5\baselineskip}

\begin{subtable}{\textwidth}
\centering
\caption{R\textsuperscript{2} Score}
\small
\resizebox{\textwidth}{!}{
\begin{tabular}{l|ccccccccccccc}
\hline
Dataset & \texttt{direct\_sr} & \texttt{vANN} & \texttt{l2} & \texttt{lip\_0.1} & \texttt{lip\_0.3} & \texttt{lip\_0.5} & \texttt{lip\_0.7} & \texttt{lip\_1} & \texttt{jac\_0.1} & \texttt{jac\_0.3} & \texttt{jac\_0.5} & \texttt{jac\_0.7} & \texttt{jac\_1} \\
\hline
California Housing & 0.427 & 0.771 & 0.771 & 0.645 & 0.555 & 0.488 & 0.419 & 0.278 & 0.628 & 0.559 & 0.509 & 0.460 & 0.425 \\
Concrete Strength & 0.097 & 0.843 & 0.835 & 0.764 & 0.648 & 0.546 & 0.462 & 0.313 & 0.733 & 0.621 & 0.575 & 0.530 & 0.480 \\
\hline
\end{tabular}
}
\end{subtable}

\vspace{4pt}
\footnotesize
\centering
\textbf{Notes.} 
\texttt{direct\_sr} = direct symbolic regression baseline;
\texttt{vANN} = vanilla ANN; 
\texttt{l2} = $\ell_2$ weight decay; 
\texttt{lip\_$\alpha$} = Lipschitz regularizer weight $\alpha$; 
\texttt{jac\_$\alpha$} = Jacobian regularizer weight $\alpha$.

\end{table*}

\textbf{Stage 1: Training the Teacher Network.}
Given a dataset $D = \{(x_i, y_i)\}_{i=1}^N$, where $x_i \in \mathbb{R}^d$ are the input features and $y_i \in \mathbb{R}$ are the target values, we first train a teacher model. The teacher is a deep neural network, $f_{\text{ANN}}(\cdot; \theta)$, with parameters $\theta$. The network is trained to minimize the standard Mean Squared Error (MSE) loss:
\begin{equation}
    \mathcal{L}_{\text{MSE}}(\theta) = \frac{1}{N} \sum_{i=1}^N (y_i - f_{\text{ANN}}(x_i; \theta))^2
    \label{eq:mse}
\end{equation}
The optimization process yields the trained teacher network, $f^*_{\text{ANN}}$.

\textbf{Stage 2: Distilling the Student Model.}
Next, we train a symbolic student model, $f_{\text{SR}}(\cdot)$, to mimic the teacher. We generate a new ``distillation'' dataset, $D' = \{(x_i, \hat{y}_i)\}_{i=1}^N$, where the targets are the predictions from the trained teacher network: $\hat{y}_i = f^*_{\text{ANN}}(x_i)$. The Symbolic Regression (SR) algorithm then searches the space of mathematical expressions, $\mathcal{F}$, to find the function $f_{\text{SR}}$ that best fits the teacher's behavior:
\begin{equation}
    f_{\text{SR}} = \arg\min_{f \in \mathcal{F}} \sum_{i=1}^N (\hat{y}_i - f(x_i))^2
    \label{eq:sr}
\end{equation}

\subsection{Jacobian Regularization (Local Smoothness)}
The function $f^*_{\text{ANN}}$ learned via the standard process can be unnecessarily complex. To address this, we investigate a regularizer based on the Jacobian matrix, $J_x f(x; \theta)$, which contains the partial derivatives of the network's output with respect to its inputs. We hypothesize that penalizing the norm of the Jacobian, a measure of local sensitivity, will encourage smoother functions. The regularized loss is:
\begin{equation}
    \mathcal{L}_{\text{Jacobian}}(\theta) = \mathcal{L}_{\text{MSE}}(\theta) + \lambda \cdot \mathbb{E}_{x \sim D_x} \left[ \|J_x f(x; \theta)\|_F^2 \right]
    \label{eq:total_loss}
\end{equation}
Here, $\| \cdot \|_F^2$ is the squared Frobenius norm. We compute the Jacobian efficiently via PyTorch's autograd engine. The expectation $\mathbb{E}_{x \sim D_x}$ is approximated by averaging over each mini-batch of size 128 during training. The primary computational cost arises from this term, which scales roughly as $\mathcal{O}(d \cdot m)$, where $d$ is the input dimension and $m$ is the network width.

\subsection{Lipschitz Regularization (Global Smoothness)}
As an alternative, we investigate a global smoothness constraint via Lipschitz regularization~\cite{gouk2021regularisation}. The Lipschitz constant of a function bounds its maximum rate of change. We approximate this by penalizing the spectral norm of the weight matrices of each linear layer in the network:
\begin{equation}
    \mathcal{L}_{\text{Lipschitz}}(\theta) = \mathcal{L}_{\text{MSE}}(\theta) + \lambda \cdot \sum_{l \in L} \|W_l\|_2
    \label{eq:lip_loss}
\end{equation}
where $L$ is the set of linear layers and $\|W_l\|_2$ is the spectral norm of the weight matrix of layer $l$. This regularizer is computationally cheaper than the Jacobian penalty.

\subsection{Theoretical Distinction: Local vs. Global}
A critical contribution of this work is determining whether symbolic distillation benefits more from local or global smoothness constraints. While both regularizers aim to simplify the learned function, they operate through fundamentally different mechanisms with distinct theoretical implications.

\textbf{Local Smoothness (Jacobian) and Non-Uniform Complexity.}
Jacobian regularization is inherently data-dependent. By penalizing $||\nabla_x f(x)||_F^2$ only at the training points $x \sim \mathcal{D}$, it forces the teacher function to be locally flat in the immediate vicinity of the data manifold. Crucially, this imposes no direct constraint on the function's behavior in empty regions of the input space. This allows the ANN to retain sharp transitions or high-frequency components if they are necessary to fit specific clusters of data, effectively permitting \textit{non-uniform complexity}. For symbolic distillation, this is conceptually advantageous: it ``cleans up'' the function where it matters (near data) without arbitrarily limiting the model's expressivity in complex regimes. However, this comes at a computational cost: calculating the full Jacobian norm requires second-order information (gradients of gradients), which scales linearly with the number of output dimensions and can be memory-intensive for wide networks.

\textbf{Global Smoothness (Lipschitz) and Uniform Constraints.}
In contrast, Lipschitz regularization is model-dependent and enforces a strict global constraint. By penalizing the spectral norm of the weight matrices ($||W||_2$), we bound the maximum possible rate of change of the function $K \approx \prod ||W_l||_2$ everywhere in the input space, regardless of the data distribution. This effectively places a ``speed limit'' on the function, ensuring it cannot oscillate wildly even in unseen regions. Theoretically, this robustness is desirable for generalization. However, in the context of distillation, a strict global Lipschitz bound might be overly conservative, forcing the teacher to underfit sharp features in the data just to satisfy the global smoothness requirement. Computationally, however, it is highly efficient, as spectral norms can be estimated quickly via power iteration without traversing the backward pass twice.

\subsection{Quantifying Functional Complexity}
To rigorously characterize the trade-off between accuracy and smoothness, we define two distinct evaluation metrics. These are computed \textit{post-hoc} on the test set to assess the properties of the trained teacher models.

\textbf{Average Jacobian Sensitivity ($S_{Jac}$).}
To measure local functional complexity, we compute the average Frobenius norm of the Jacobian matrix over the test set $D_{test}$:
\begin{equation}
S_{Jac}(f) = \frac{1}{|D_{test}|} \sum_{x \in D_{test}} ||J_x f(x)||_F
\end{equation}
A lower $S_{Jac}$ indicates that the function is locally smooth and insensitive to small perturbations in the input, a property we hypothesize correlates with distillability.

\textbf{Lipschitz Constant Approximation ($S_{Lip}$).}
To measure global complexity, we estimate (via an upper bound proxy) the network's Lipschitz constant by computing the product of the spectral norms of the weight matrices for all layers $l=1 \dots L$:
\begin{equation}
S_{Lip}(f) = \prod_{l=1}^{L} ||W_l||_2
\end{equation}
where $||W_l||_2$ is the largest singular value of the weight matrix $W_l$. A lower $S_{Lip}$ implies a globally slower-changing function with bounded gradients.

By mapping the teachers on the 2D plane of Accuracy ($R^2$) versus Complexity ($S_{Jac}$ or $S_{Lip}$), we can identify the Pareto frontier of models that balance predictive power with interpretability.






\begin{figure}[!t]
    \centering
    \includegraphics[width=\columnwidth]{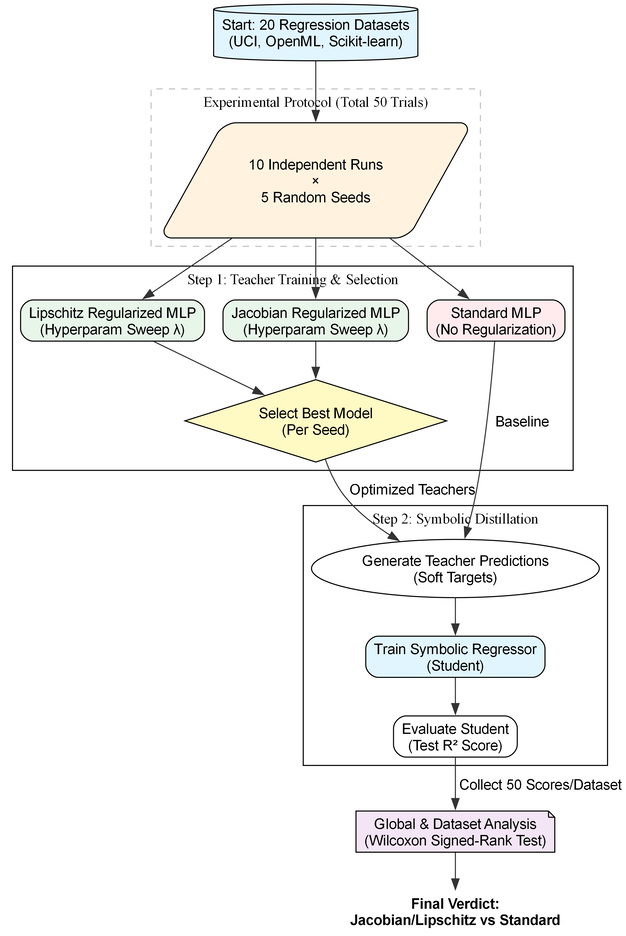}
    \caption{Main Distillation Experiment Pipeline}
    \label{fig:dist1}
\end{figure}

\section{Experimental Results and Analysis}
\label{sec:experiments}

Our experiments are designed to rigorously compare a standard distillation baseline against Jacobian (local) and Lipschitz (global) smoothness regularizers. We investigate their impact when distilling to Symbolic Regression (SR) students with varying expressive power. The experimental pipeline flowchart for the main distillation experiment is shown in Figure \ref{fig:dist1}.

\subsection{The Accuracy-Smoothness Problem}
\begin{figure*}[!h]
    \centering

    \includegraphics[width=0.64\textwidth]{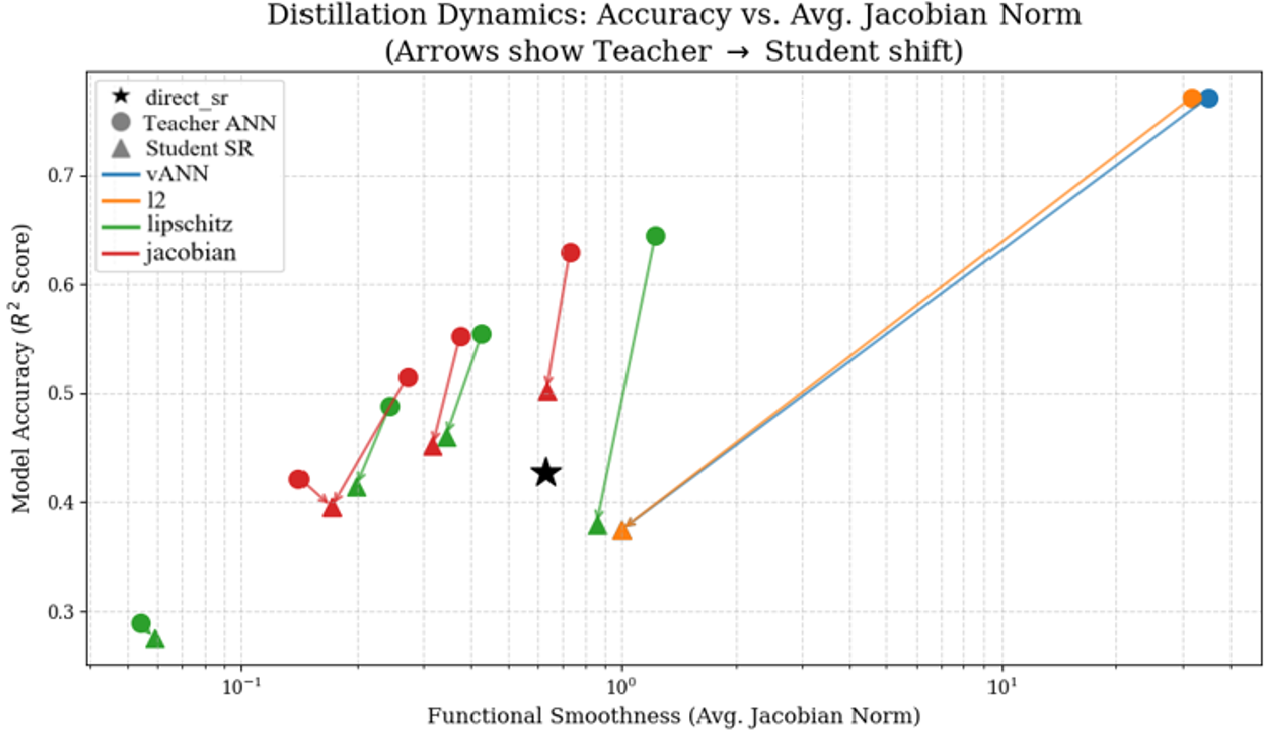}\vspace{1em}
    \includegraphics[width=0.64\textwidth]{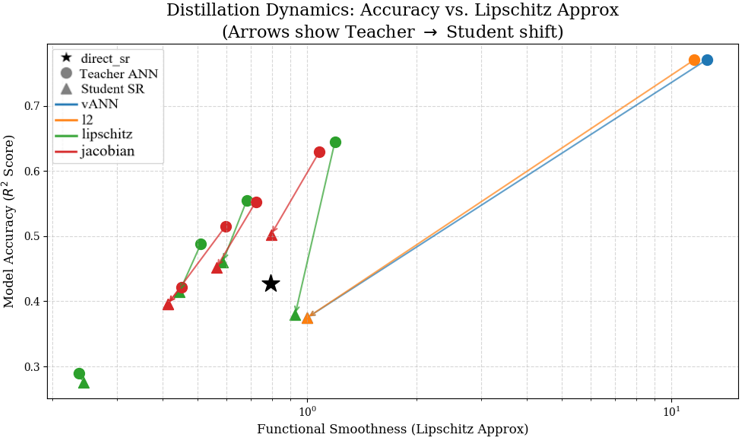}
    \vspace{0.5em}

    {\footnotesize
    \texttt{direct\_sr} = direct symbolic regression baseline;
    \texttt{vANN} = vanilla ANN;
    \texttt{l2} = $\ell_2$ weight decay;\\
    \texttt{lipschitz} = Lipschitz regularizer weight $\alpha$;
    \texttt{jacobian} = Jacobian regularizer weight $\alpha$.
    \par}
    
    \caption{\textbf{Accuracy vs. Functional Smoothness Trade-offs.}}
    \textbf{Top:} Jacobian-regularized models. 
    \textbf{Bottom:} Lipschitz-regularized models.
    
    \textbf{Interpretation:} This visualization characterizes the distillation landscape. The \textbf{$x$-axis} represents functional complexity (log scale, lower is smoother), while the \textbf{$y$-axis} represents predictive accuracy ($R^2$). 
    \textbf{(1) The Extremes:} Standard ANNs (blue/orange) occupy the top-right (accurate but ``spiky''), whereas \texttt{direct\_sr} (black star) sits in the lower-left (smooth but inaccurate). 
    \textbf{(2) The ``Sweet Spot'':} Regularized teachers (red/green) trace a trajectory between these extremes. An optimal teacher is found where the curve passes \textit{above and to the left} of the \texttt{direct\_sr} baseline, indicating a model that is simultaneously more accurate and smoother than the symbolic baseline. 
    \textbf{(3) Distillation Gap:} Arrows point from Teacher ($\bullet$) to Student ($\blacktriangle$). Shorter arrows indicate successful knowledge transfer; note how students of regularized teachers maintain accuracy significantly better than students of standard ANNs.
    \label{fig:gp}
\end{figure*}

We first characterize the problem. We plot all models on a 2D plane of Accuracy ($R^2$) vs.\ Smoothness (e.g., Avg.\ Jac.\ Norm) in Figure \ref{fig:gp} using \texttt{gplearn}. We observe two extremes: standard ANNs (\texttt{vANN}, \texttt{l2}) occupy the top-right (accurate but spiky), while \texttt{direct\_sr} occupies the somewhat bottom-left (smoother but inaccurate). This illustrates the fundamental challenge. Our aim is to find a ``sweet spot" teacher ANN that will be more accurate and smoother than \texttt{direct\_sr}.

\subsection{Finding the ``Sweet Spot" Teacher}
Our regularized models (\texttt{jacobian}, \texttt{lipschitz}) form a trade-off curve connecting these extremes. As $\lambda$ increases, the models trace a path from the spiky \texttt{vANN} baseline toward the \texttt{direct\_sr} baseline and beyond. This allows us to find a ``sweet spot.'' For example in Table \ref{tab:combined_results}, for California Housing, \texttt{jacobian\_0.3} ($R^2 = 0.56$, Jac. Norm $= 0.39$) is superior to \texttt{direct\_sr} ($R^2 = 0.43$, Jac. Norm $= 0.63$) on both metrics. We have successfully created an ANN that is both more accurate and smoother than the direct symbolic solution.

\subsection{Main Distillation Experiment}

We hypothesize that these ``sweet spot'' teachers will produce superior student models. To evaluate the robustness of our proposed regularization methods, we conducted a large-scale empirical study across 20 diverse regression datasets sourced from the UCI Machine Learning Repository, OpenML, and Scikit-learn. All datasets used in our experiments are publicly available. 


\subsubsection{Experimental Design}
To account for the stochastic nature of both Neural Network initialization and Genetic Programming evolution, we implemented a rigorous experimental protocol with a total of 50 independent trials per dataset. Specifically, we executed 10 independent experimental runs, with each run averaging results over 5 distinct random seeds ($10 \text{ runs} \times 5 \text{ seeds} = 50 \text{ trials}$).

For each seed, we trained a standard unregularized Teacher (MLP) and two regularized variants: Jacobian regularization (local smoothness) and Lipschitz regularization (global smoothness). We performed a hyperparameter sweep for the regularization strength $\lambda \in \{0.001, 0.005, 0.01, 0.05, 0.1, 0.2, 0.5, 1.0\}$ and selected the best-performing model on the train set for each method per seed. All statistical comparisons were performed using the Wilcoxon Signed-Rank Test, a non-parametric paired difference test suitable for comparing machine learning algorithms across multiple datasets.

\subsubsection{Global Statistical Analysis}
Our primary objective was to determine if smoothness-regularized teachers consistently produce superior symbolic students compared to standard teachers. We aggregated the mean $R^2$ scores for each method across all 20 datasets and performed a global Wilcoxon Signed-Rank Test.

As shown in Table~\ref{tab:global_stats}, both regularization methods achieved statistically significant improvements over the standard baseline.
\begin{itemize}
    \item \textbf{Jacobian vs. Standard:} The Jacobian regularizer yielded a mean $R^2$ improvement of \textbf{0.0531} ($p = 4.77 \times 10^{-7}$), demonstrating a highly significant advantage.
    \item \textbf{Lipschitz vs. Standard:} The Lipschitz regularizer also significantly outperformed the baseline, with a mean $R^2$ improvement of \textbf{0.0364} ($p = 4.77 \times 10^{-7}$).
\end{itemize}

\begin{table}[!t]
    \centering
    \caption{\textbf{Global Statistical Analysis (Wilcoxon Signed-Rank Test).} Comparison of mean $R^2$ improvement across all 20 datasets. Both regularizers significantly outperform the Standard baseline, with Jacobian showing a slight edge over Lipschitz.}
    \label{tab:global_stats}
    \resizebox{\columnwidth}{!}{%
    \begin{tabular}{lcc}
        \toprule
        \textbf{Comparison} & \textbf{Mean Diff. ($R^2$)} & \textbf{\textit{p}-Value} \\
        \midrule
        Jacobian vs. Standard & $+0.0531$ & $4.77 \times 10^{-7}$ \\
        Lipschitz vs. Standard & $+0.0364$ & $4.77 \times 10^{-7}$ \\
        Jacobian vs. Lipschitz & $+0.0168$ & $2.85 \times 10^{-3}$ \\
        \bottomrule
    \end{tabular}%
    }
\end{table}

Furthermore, a direct comparison between the two regularizers revealed that Jacobian regularization provided a statistically significant edge over Lipschitz regularization ($p = 2.85 \times 10^{-3}$), suggesting that enforcing local smoothness may be slightly more beneficial for distillability than global spectral constraints on this specific suite of regression tasks.

\subsubsection{Dataset-Level Analysis}
We further analyzed performance at the individual dataset level. As detailed in Table~\ref{tab:dataset_stats}, where we show a subset of the full results, the regularized teachers achieved a statistically significant improvement (at $p < 0.05$) against the standard teacher.

\begin{table}[!t]
    \centering
    \caption{\textbf{Selected Dataset-Level Performance.} Comparison of the student model's test $R^2$ score (distilled from Jacobian/Lipschitz teachers) versus the Standard baseline. Statistical significance is assessed via \textit{p}-values ($p < 0.05$). See Table \ref{tab:full_results} for full results on all 20 datasets.}
    \label{tab:dataset_stats}
    \resizebox{\columnwidth}{!}{%
    \begin{tabular}{lccc}
        \toprule
        \textbf{Dataset} & \textbf{Method} & \textbf{Mean $R^2$} & \textbf{\textit{p}-Val vs Std} \\
        \midrule
        \multirow{2}{*}{Electrical Grid Stability} & Jac & \textbf{0.6814} & 0.0020 \\
         & Lip & 0.6778 & 0.0039 \\
        \midrule
        \multirow{2}{*}{Auto MPG} & Jac & \textbf{0.7615} & 0.0020 \\
         & Lip & 0.7576 & 0.0039 \\
        \midrule
        \multirow{2}{*}{Combined Cycle Power} & Jac & \textbf{0.9105} & 0.0234 \\
         & Lip & 0.9093 & 0.0391 \\
        \midrule
        \multirow{2}{*}{fifa} & Jac & \textbf{0.3349} & 0.0020 \\
         & Lip & 0.3245 & 0.0020 \\
        \midrule
        \multirow{2}{*}{mu284} & Jac & \textbf{0.6733} & 0.0039 \\
         & Lip & 0.5917 & 0.0039 \\
        \bottomrule
    \end{tabular}%
    }
\end{table}

\begin{table*}[t]
    \centering
    \caption{\textbf{Full Experimental Results across 20 Datasets.} Comparison of Student $R^2$ scores distilled from Standard (Unregularized) teachers versus Jacobian- and Lipschitz-regularized teachers. The ``Standard'' column reports baseline performance. The statistical significance of the improvement in $R^2$ is assessed via one-sided paired tests against the baseline.}
    \label{tab:full_results}
    \begin{tabular}{l|c|cc|cc}
        \toprule
        \multirow{2}{*}{\textbf{Dataset}} 
        & \textbf{Standard Pipeline (Baseline)} 
        & \multicolumn{2}{c|}{\textbf{Jacobian Pipeline}} 
        & \multicolumn{2}{c}{\textbf{Lipschitz Pipeline}} \\
         & \textbf{Mean $R^2$} 
         & \textbf{Mean $R^2$} & \textbf{\textit{p}-val} 
         & \textbf{Mean $R^2$} & \textbf{\textit{p}-val} \\
        \midrule
        Airfoil Self-Noise & 0.3446 & \textbf{0.3872} & 0.006 & 0.3806 & 0.004 \\
        Elec. Grid Stability & 0.6559 & \textbf{0.6814} & 0.002 & 0.6778 & 0.004 \\
        Real Estate Val. & 0.4782 & \textbf{0.4992} & 0.002 & 0.4962 & 0.010 \\
        AI4I 2020 Maintenance & 0.5391 & \textbf{0.6255} & 0.039 & 0.5395 & 0.063 \\
        Auto MPG & 0.7493 & \textbf{0.7615} & 0.002 & 0.7576 & 0.004 \\
        Combined Cycle Power & 0.9090 & \textbf{0.9105} & 0.023 & 0.9093 & 0.039 \\
        Cholesterol & 0.0070 & \textbf{0.2138} & 0.002 & 0.0661 & 0.002 \\
        Munich Rent Index & 0.0994 & \textbf{0.1400} & 0.002 & 0.1222 & 0.002 \\
        SEA (50000) & 0.3902 & \textbf{0.3985} & 0.002 & 0.3980 & 0.010 \\
        kin8nm & 0.2913 & \textbf{0.3024} & 0.008 & 0.3008 & 0.008 \\
        Liver Disorders & 0.0533 & \textbf{0.0877} & 0.020 & 0.0823 & 0.004 \\
        Brazilian Houses & 0.0140 & \textbf{0.0408} & 0.002 & 0.0402 & 0.002 \\
        Miami Housing 2016 & 0.3957 & 0.4159 & 0.012 & \textbf{0.4227} & 0.002 \\
        Elevators & 0.3341 & \textbf{0.3495} & 0.004 & 0.3494 & 0.004 \\
        FIFA & 0.1114 & \textbf{0.3349} & 0.002 & 0.3245 & 0.002 \\
        analcatdata\_apnea3 & 0.2015 & \textbf{0.2722} & 0.004 & 0.2650 & 0.002 \\
        mu284 & 0.4367 & \textbf{0.6733} & 0.004 & 0.5917 & 0.004 \\
        Places & 0.0457 & \textbf{0.0765} & 0.002 & 0.0695 & 0.004 \\
        Diabetes & 0.3695 & \textbf{0.3993} & 0.004 & 0.3889 & 0.004 \\
        Friedman \#1 & 0.5667 & \textbf{0.5927} & 0.020 & 0.5897 & 0.020 \\
        \bottomrule
    \end{tabular}
\end{table*}

\begin{figure}[!t]
    \centering
    \includegraphics[width=\columnwidth]{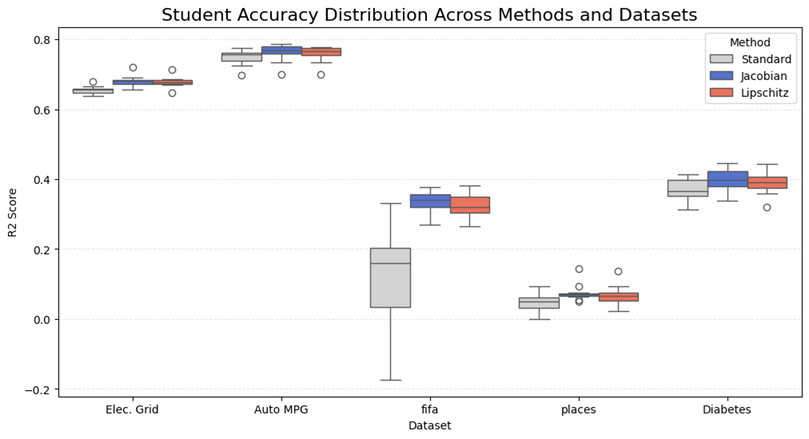}
    \caption{Distribution of Student Accuracy: Box Plot}
    \label{fig:box}
    \vspace{-4mm}
\end{figure}

\begin{table}[h]
\centering
\caption{Key Hyperparameters for Experimental Models}
\label{tab:app_hyperparams}
\begin{tabular}{l p{0.6\linewidth}}
\toprule
\textbf{Model / Parameter} & \textbf{Value} \\
\midrule
\multicolumn{2}{l}{\textit{\textbf{ANN Teacher (PyTorch)}}} \\
Architecture       & 2 hidden layers (100 neurons each) \\
Activation Function& ReLU \\
Optimizer          & Adam \\
Learning Rate      & 0.001 \\
Epochs             & 200 \\
Batch Size         & 128 \\
\midrule
\multicolumn{2}{l}{\textit{\textbf{Symbolic Regressor Student (\texttt{gplearn})}}} \\
Population Size    & 1000 \\
Generations        & 10 \\
Metric             & Mean Squared Error (MSE) \\
Function Set       & \{+, -, *, /\} \\
Random State       & 50 Independent Seeds ($10 \text{ runs} \times 5 \text{ seeds}$) \\
\bottomrule
\end{tabular}
\end{table}

\begin{figure}[!t]
    \centering
    \includegraphics[width=0.9\columnwidth]{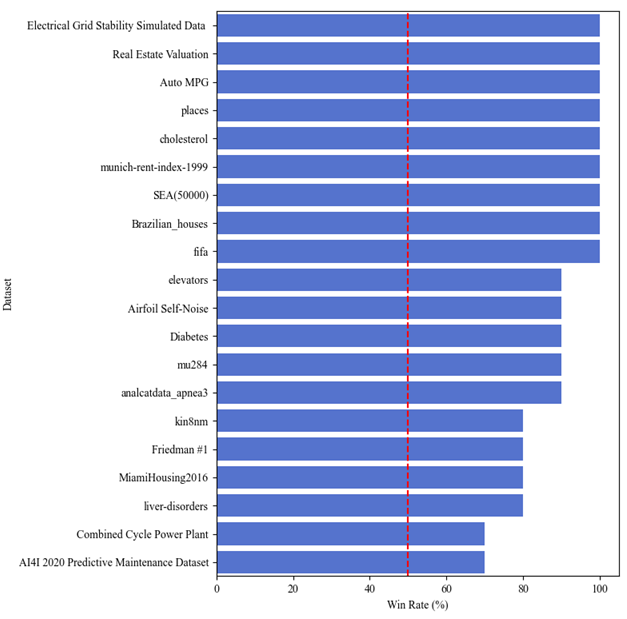}\vspace{-2mm}
    \caption{Consistency Chart: Percentage of Runs where Jacobian outperformed Standard}
    \label{fig:sub}\vspace{-5mm}
\end{figure}

These results confirm that active smoothness regularization is a robust strategy for improving symbolic distillation, consistently yielding students that are as good as or significantly better than those distilled from standard networks. Figure \ref{fig:box} reveals that for datasets like \texttt{Electrical Grid Stability} and \texttt{Auto MPG}, the Jacobian-regularized students consistently outperform the baseline, with the entire interquartile range shifted upwards. Furthermore, Figure \ref{fig:sub} demonstrates the robustness of the proposed approach, with the Jacobian-regularized pipeline outperforming the standard teacher in between 70\% and 100\% of independent runs, depending on the dataset.

\section{Discussion}

Our experiments demonstrate that functional smoothness regularization is a powerful, robust tool for improving symbolic distillation. By scaling our evaluation to 20 datasets and 50 independent trials, we have moved beyond anecdotal evidence to statistically validate the ``Sweet Spot'' hypothesis. Table~\ref{tab:full_results} provides a summary of the full experimental results used in our final evaluation. The hyperparameters for the teacher and student models were standardized across all experiments to ensure a fair comparison. The key parameters, derived from our Python implementation, are detailed in Table~\ref{tab:app_hyperparams}.

\subsection{The Accuracy-Smoothness Trade-off}
The core challenge of symbolic distillation remains the fundamental misalignment in complexity. Standard ANNs are highly accurate but functionally irregular (``spiky''), while direct Symbolic Regression is smooth but often inaccurate. Our broad empirical study confirms that this is not an isolated phenomenon but a pervasive issue across diverse regression tasks. The regularized teachers consistently bridge this gap, tracing a trade-off curve that allows us to purchase ``distillability'' (smoothness) at a small, controlled cost to the teacher's raw accuracy.

\subsection{Validating the ``Sweet Spot'' Teacher}
The solution is to tune the teacher to an optimal ``sweet spot'' where it is simple enough to be distilled but accurate enough to be useful. In our expanded experiments, the effectiveness of this approach is evident.

Crucially, the Global Statistical Analysis confirms that these gains are not due to random seed variation. The Wilcoxon Signed-Rank Test results ($p < 10^{-6}$) provide strong evidence that active smoothness regularization systematically improves the ``teachability'' of neural networks, making it a reliable default strategy for distillation pipelines.

\subsection{Local vs. Global Smoothness}
A novel finding from this large-scale study is the comparative performance of the regularizers. While both methods significantly beat the standard baseline, the Jacobian regularizer (local smoothness) demonstrated a statistically significant advantage over the Lipschitz regularizer (global smoothness) in our global analysis ($p \approx 0.0028$).

This suggests that for tabular regression tasks, constraining the \textit{local} sensitivity of the model (via the Jacobian) might preserve more useful information than enforcing a strict \textit{global} spectral bound (via Lipschitz). However, this comes with a trade-off in scalability: Jacobian regularization requires calculating second-order information (double backpropagation), which scales poorly with input dimension compared to the computationally cheaper spectral normalization of the Lipschitz approach.


\subsection{Ablation Study}

Our primary distillation experiments in Section \ref{sec:experiments} utilized \texttt{gplearn}, a standard Genetic Programming-based symbolic regressor, to ensure our findings were not dependent on a specific high-end optimization algorithm. However, to verify that the ``smoothness hypothesis'' is fundamental to the distillation process rather than an artifact of a specific SR implementation, we conducted an ablation study using \texttt{PySR} \cite{13cranmer2023interpretablemachinelearningscience}, a state-of-the-art SR framework that typically produces higher-quality and more compact formulas than \texttt{gplearn}.

\begin{figure}[!t]
    \centering
    \includegraphics[width=\columnwidth]{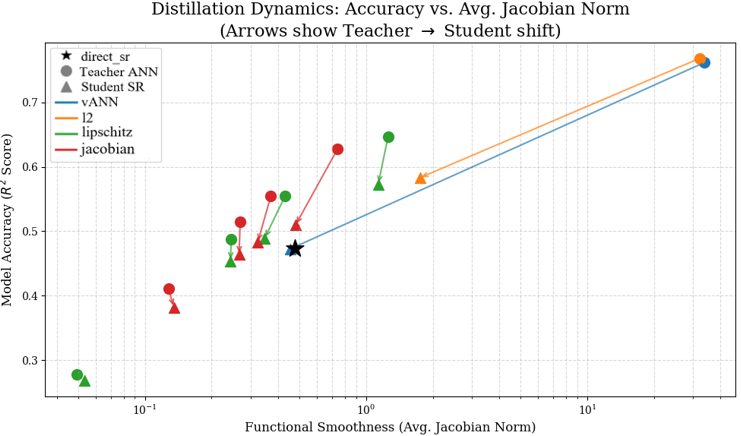}\vspace{1em}
    \includegraphics[width=\columnwidth]{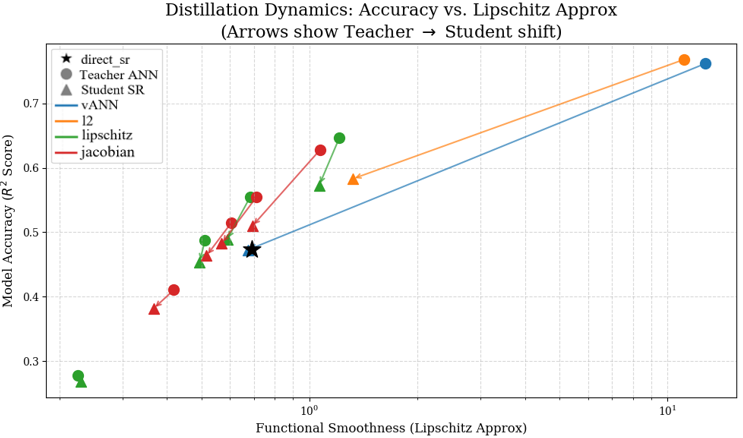}

    {\footnotesize
    \texttt{direct\_SR} = direct symbolic regression baseline;
    \texttt{vANN} = vanilla ANN;
    \texttt{l2} = $\ell_2$ weight decay;
    \texttt{lipschitz} = Lipschitz regularizer weight $\alpha$;
    \texttt{jacobian} = Jacobian regularizer weight $\alpha$.
    }
    \caption{Accuracy vs functional smoothness trade-offs on the California housing dataset using \texttt{PySR}.
    \textbf{Top:} Jacobian-regularized models.
    \textbf{Bottom:} Lipschitz-regularized models.}
    \label{fig:pysr}
    \vspace{-10mm}
\end{figure}

We replicated the accuracy--smoothness trade-off characterization on the California Housing dataset using PySR. As shown in Figure~\ref{fig:pysr}, the same dynamics consistently emerge across multiple random initializations:
\begin{enumerate}
    \item \textbf{The Mismatch Persists:} Even with PySR's advanced search capabilities, the standard ANN teacher remains in the ``high-accuracy, high-complexity'' regime, while the direct symbolic regression baseline remains smooth but inaccurate.
    \item \textbf{Universal Benefit:} The smoothness-regularized teachers (Jacobian and Lipschitz) again bridge this gap, tracing a Pareto frontier that allows PySR to recover significantly more accurate symbolic models compared to the direct baseline.
\end{enumerate}

This confirms that functional smoothness is a prerequisite for effective symbolic distillation and compact symbolic representations, regardless of whether the student model is a traditional GP (\texttt{gplearn}) or a modern, high-performance symbolic search algorithm (\texttt{PySR}).

\subsection{Limitations and Future Work}

While robust, our method is not a panacea. First, this work focuses exclusively on Artificial Neural Networks (ANNs) as teachers. While knowledge distillation is applicable to other black-box models (e.g., Gradient Boosting Machines or Random Forests), ANNs represent the overwhelmingly dominant teacher architecture in current research and industrial practice. Extending this smoothness-regularization framework to other classes of teacher models, or utilizing GPU-accelerated evolutionary strategies \cite{21wang2024tensorizedneuroevolutionaugmentingtopologies} to scale the symbolic search, remains an open and practically important avenue for investigation.

Second, our experiments indicate that the ``sweet spot'' is dataset-dependent. Future work should focus on adaptive regularization schemes that can automatically tune $\lambda$ during training, thereby removing the need for a costly hyperparameter grid search. Additionally, exploring this framework on higher-dimensional modalities, such as time-series or image data, remains an exciting and necessary path forward for advancing robust and interpretable distillation \cite{23ross2018improving, 27raghu2020surveydeeplearningscientific}.

\section{Conclusion}

We identified that the failure of symbolic distillation stems from a fundamental misalignment: standard ANNs are too functionally complex (``spiky'') to be effective teachers, while direct Symbolic Regression is often too inaccurate. We solve this by introducing functional smoothness regularizers, not merely to constrain the model, but to actively tune a teacher to a ``sweet spot'' that balances accuracy and simplicity.

We rigorously validated this framework through an extensive empirical study across 20 diverse regression datasets, conducting 50 independent trials per dataset. Our results are unequivocal: global statistical analysis using the Wilcoxon Signed-Rank Test confirms that students distilled from smoothness-regularized teachers achieve significantly higher accuracy ($p < 10^{-6}$) than those distilled from standard networks. Furthermore, our analysis reveals that Jacobian regularization (local smoothness) provides a slight but statistically significant edge over Lipschitz regularization (global smoothness) in this domain. By establishing functional smoothness as a critical prerequisite for distillability, this work provides a principled path toward extracting reliable, transparent, and trustworthy equations from black-box AI systems.

\section*{ACKNOWLEDGMENTS}
This research/project is supported by the National Research Foundation, Singapore under its AI Singapore Programme (AISG Award No: AISG3-PhD-2023-08-052T), A*STAR, CISCO Systems (USA) Pte. Ltd and National University of Singapore under its Cisco-NUS Accelerated Digital Economy Corporate Laboratory (Award I21001E0002), and the Ministry of Education, Singapore, under its Academic Research Fund Tier 2 (Award No.: MOE-T2EP20125-0013).

\bibliographystyle{ACM-Reference-Format}
\bibliography{sample-base}

\end{document}